\documentclass[lettersize,journal]{IEEEtran}
\usepackage{amsmath,amsfonts,amssymb,bm}
\usepackage{algorithmic}
\usepackage{algorithm}
\usepackage{array}
\usepackage[caption=false,font=normalsize,labelfont=sf,textfont=sf]{subfig}
\usepackage{textcomp}
\usepackage{stfloats}
\usepackage{url}
\usepackage{verbatim}
\usepackage{graphicx}
\usepackage{cite}
\usepackage{epstopdf}
\usepackage{hyperref}
\usepackage{times}
\usepackage{CJK}
\usepackage{latexsym}
\usepackage{multicol}
\usepackage{multirow}
\usepackage{booktabs}
\usepackage{arydshln}
\usepackage{xcolor}
\usepackage[T1]{fontenc}
\usepackage{tikz}

\definecolor{lime}{HTML}{A6CE39}
\DeclareRobustCommand{\orcidicon}{%
    \begin{tikzpicture}
    \draw[lime, fill=lime] (0,0) circle [radius=0.16] 
    node[white] {{\fontfamily{qag}\selectfont \tiny ID}};
    \draw[white, fill=white] (-0.0625,0.095) circle [radius=0.007];
    \end{tikzpicture}\hspace{-2mm}}

\foreach \x in {A, ..., Z}{%
    \expandafter\xdef\csname orcid\x\endcsname{%
      \noexpand\href{https://orcid.org/\csname orcidauthor\x\endcsname}{\noexpand\orcidicon}}%
}

\usepackage{balance}

\begin{document}

\title{AxBERT: An Interpretable Chinese Spelling Correction Method Driven by Associative Knowledge Network}

\author{Fanyu Wang\orcidA{}, \emph{Student Member, IEEE}, Hangyu Zhu, and Zhenping Xie\orcidB{} \emph{Member, IEEE}%
\thanks{This work was supported in part by the National Natural Science Foundation of China (NSFC) under Grant 62272201 and 61872166; in part by the Six Talent Peaks Project of Jiangsu Province under Grant 2019 XYDXX-161.}%
\thanks{The authors are with the School of Artificial Intelligence and Computer Science and the Jiangsu Key Laboratory of Media Design and Software Technology, Jiangnan University, Wuxi 214122, Jiangsu, China (e-mail: xiezp@jiangnan.edu.cn).}%
}

\maketitle

\begin{abstract}
Deep learning has shown promising performance on various machine learning tasks. Nevertheless, the uninterpretability of deep learning models severely restricts the usage domains that require feature explanations, such as text correction. Therefore, a novel interpretable deep learning model (named AxBERT) is proposed for Chinese spelling correction by aligning with an associative knowledge network (AKN). Wherein AKN is constructed based on the co-occurrence relations among Chinese characters, which denotes the interpretable statistic logic contrasted with uninterpretable BERT logic. And a translator matrix between BERT and AKN is introduced for the alignment and regulation of the attention component in BERT. In addition, a weight regulator is designed to adjust the attention distributions in BERT to appropriately model the sentence semantics. Experimental results on SIGHAN datasets demonstrate that AxBERT can achieve extraordinary performance, especially upon model precision compared to baselines. Our interpretable analysis, together with qualitative reasoning, can effectively illustrate the interpretability of AxBERT.\footnote{The all code and dataset for this paper is available.}
\end{abstract}


\section{Introduction}

\label{related_work}
\IEEEPARstart{T}{ext} correction methods serve as essential tools for people in various application scenarios, such as machine translation, office writing assistance, etc \cite{ghufron2018role,napoles-etal-2017-jfleg,omelianchuk-etal-2020-gector}. Wherein with the development of search engines, speech recognition, and so on, spelling correction is currently the most commonly used text correction method, which aims to optimize the inputting text and improve the prediction performance of the whole framework.

\emph{Interpretability}, much-needed for AI, describes the capacity of the methods to explain the basis of the decision to people \cite{mittelstadt2019explaining,beckh2021explainable}. Regardless of the extraordinary performance achieved by the recent spelling correction methods, the increasing uninterpretability constrains the further application of the methods in some specific domains \cite{Holzinger2018,seeliger2019semantic} and adversely affect the trust of AI methods\cite{gunning2019darpa}.

The fact is that \emph{no interpretable method exists for spelling correction with a good performance}. The rule-based correction methods are widely employed in specific domains required for controllable correction, such as medical and biological \cite{crowell2004frequency,lai2015automated}. Generally, the rule-based methods conduct the presetting rules based on the character correlations in the correction to realize the interpretable decision-making process \cite{xiong2015hanspeller,yeh2014chinese}. However, the complex semantic context determines that the complete coverage of error cases for the presetting rules is impossible, indicating that rule-based methods are insufficient to handle complex errors.
    
In recent works, the proposed transformer-based language models significantly improve the performance of spelling correction. Researchers attempt to explain the language model by use of the extraction for the character relations, but \emph{the extracted relations are irregular compared to linguistics and experience}. The irregular relations among the characters in BERT exist for two reasons:
\begin{itemize}    
    \item \emph{The redundancy is widely distributed in the extracted information from attention layers}. While attention is considered as the key component in transformer-based models \cite{chefer2021transformer, vig2019multiscale}, the function of every attention head is still undetermined. Previous researches demonstrate that the pruning of the redundant attention heads is able to improve the performance of the language model \cite{voita-etal-2019-analyzing,behnke2020losing,wang2021spatten}, which indicates that the redundancy disrupts the modeling process and severely constrains the further analysis.
    \item \emph{The layers perform different functions in transformer-based language model (BERT)}. The layer-wise analysis is conducted in BERT, and the results illustrate that the containing layers serve different functions \cite{van2019how,lu2021advances,reif2019visualizing}. Nevertheless, the existence of various layer functions challenging the comprehensive explaining of BERT, which needs a unification for transferring the inside logic to understandable logic for human beings.
\end{itemize}

\begin{figure*}[htbp]
	\centering
	\begin{minipage}{0.49\linewidth}
		\centering
        \includegraphics[width=\linewidth]{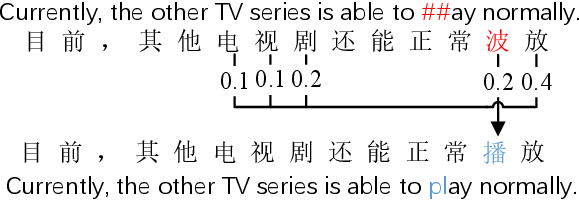}
        \caption{Example of the character relation}
        \label{fig:1}
	\end{minipage}
	\begin{minipage}{0.49\linewidth}
		\centering
        \includegraphics[width=\linewidth]{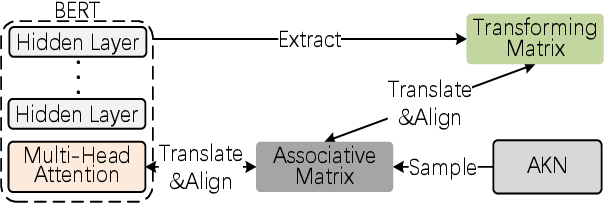}
        \caption{Information flow in AxBERT}
        \label{fig:2}
	\end{minipage}
\end{figure*}

In this work, we address the obstacles by (1) developing an interpretable method for controllable Chinese spelling correction and (2) constructing an alignment and regulation closed-loop circuit (illustrated in Figure \ref{fig:1}) to integrate the interpretable statistic logic into BERT which effectively reduce the redundancy and unify the logic among the layers. Wherein the alignment and regulation are not stated explicitly, but the interpretability of the entire framework is increased respect for the specific interpretable component with integrated information, which is demonstrated for effectiveness in previous work \cite{liu-etal-2019-knowledge,chen2020concept,Rybakov2020}. Our approach - AxBERT - consist of an \textbf{A}ssociative knowledge network (AKN, \cite{li2022associative} and (\textbf{x}) \textbf{BERT} \cite{devlin-etal-2019-bert} to realize an interpretable Chinese spelling correction method with extraordinary performance. More specifically, our main contributions are concluded as:
\begin{itemize}
    \item An \emph{associative matrix} sampled from \emph{associative knowledge network} is introduced to AxBERT, which reflected interpretable statistic logic with context localization. (section \ref{akn})
    \item We use the least-squares function to quantify transformation process in BERT as a \emph{transforming matrix} reflected as uninterpretable BERT logic. (section \ref{transforming matrix})
    \item A \emph{translator matrix} is introduced to bridge the gap between statistic logic and BERT logic, which is multiplied with the attention component in BERT to align with statistical distribution. (section \ref{translator matrix} and \ref{attention align})
    \item We introduce \emph{weight regulator}, relied on the character-level similarity between attention and AKN, to regulate the attention distributions for a appropriate semantic modeling process and better correction performance. (section \ref{weight regulator})
    \item The outstanding correction performance on SIGHAN datasets demonstrates the effectiveness of AxBERT. In Addition, the interpretable analysis is designed to exhibit the regulating process and quantitatively verify the interpretability of our proposed method (section \ref{experiments}).
\end{itemize}

\section{Related Work}
\subsection{Interpretable Analysis in BERT}
The backbones of the BERT mainly consist of uninterpretable feature representations. The existing analysis to explain BERT conduct different aspect including self-attention, linguistic knowledge, etc \cite{rogers2021}. Researchers attempt to obtain the semantic relation among the tokens to reveal the basis of the modeling result \cite{htut2019attention,goldberg2019assessing,hewitt-manning-2019-structural} (e.g., Figure \ref{fig:2}). The analysis from the attention perspective exhibit the visualization of the attention distribution and statistical results to quantitatively explain BERT \cite{clark2019does,vig2019analyzing,kovaleva2019revealing,bian-etal-2021-attention}. From the comprehensive perspective, the overrated of self-attention in the analysis is one-sided, and also disadvantages for further analysis \cite{li-etal-2019-word,Pande_Budhraja_Nema_Kumar_Khapra_2021}, because the representation transformation in BERT contains the processing from different layers. Inspired by the previous works in interpretable analysis, we comprehensively quantify transformation logic from all the layers in BERT while opting for the attention distribution as the key to integrating the interpretable statistic logic into BERT.

\subsection{Transformer-based Chinese Spelling Correction}
Benefits from the proposed transformer network, the transformer-based language models can efficiently capture the semantic information of the given sentences \cite{zhang-etal-2019-ernie, devlin-etal-2019-bert, clark2019electra}. Based on the extraordinary semantic modeling ability, the transformer-based language models are introduced for spelling correction task, which significantly enhances the correction performance. One of the principle approaches is to consider the masked token prediction task and conduct this task as correcting the inappropriate characters \cite{zhang2020spelling, cui2020revisiting}. Besides, the researchers make use of the external features of the characters, such as phonic and shape, to expand the embedding dimensions to achieve reliable correction results \cite{hong2019faspell, cheng2020spellgcn, liu2022visual}. Furthermore, the correction methods with well-performed decoders have emerged recently, which enables the candidate distributions with precise result \cite{bao-etal-2020-chunk,li2021tail}.

\subsection{Associative Knowledge Network}
\label{associative}
Associative knowledge network \cite{li2022associative}, a statistical network based on the co-occurrence among the phrases. We introduce a modified AKN to AxBERT, constructed at character level, to fit the embedding layer of BERT. For the sentences in reference articles, we initial and update AKN in AxBERT, which consist of the element $\bm{A}_{i,j}$, according to:
\begin{equation}
    \boldsymbol{A}_{i,j}=\prod_{\mathrm{sent}}\mathrm{SR}\sum_{\mathrm{sent}} \frac{1}{distance_{\langle i,j \rangle}}
    \label{eq:1}
\end{equation}
where $\bm{A} \in {\mathbb{R}^{v\times v}}$, $v$ is the length of the character list. $distance_{\langle i,j \rangle}$ is the character distance between $i-th$ character and $j-th$ character. The shorter the distance between the characters, the stronger associative relation among the characters. We find that the scores of commonly used characters will accumulate to extremely high in the updating process, the shrink rate $\mathrm{SR}$ is introduced to keep balance, which is default as $0.95$.

\begin{figure*}[!t]
    \begin{center}
        \includegraphics[width=6in]{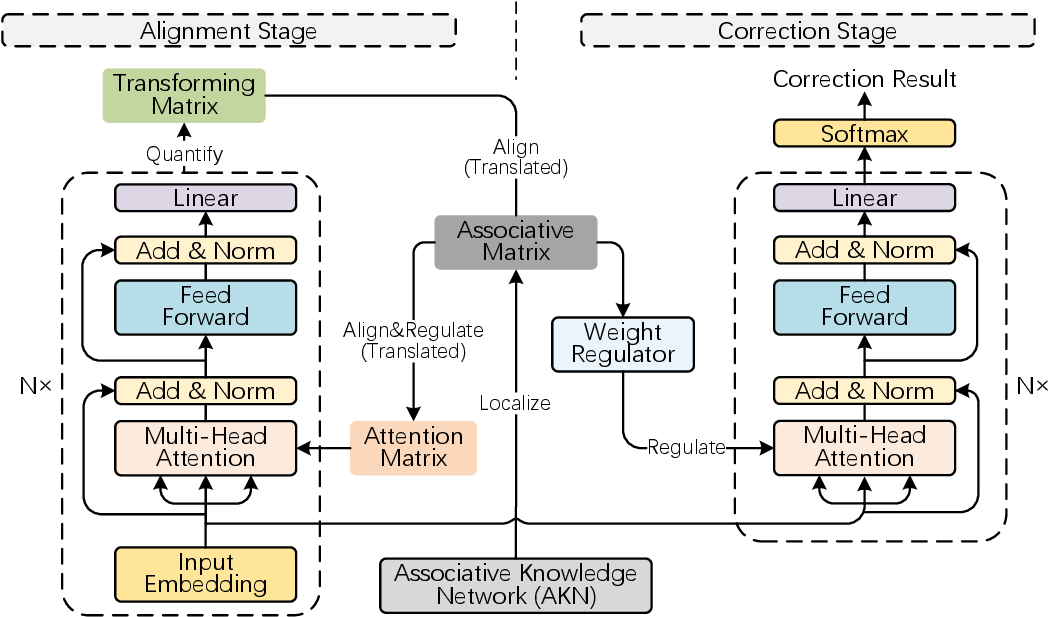}

    \end{center}
    \caption{Overview of AxBERT's architecture. The inputting given sentence is firstly modeled by BERT in the alignment stage (Left), we quantify the transformation process of the sentence representations in BERT as the transforming matrix $\bm{M}_{T}$. (Middle) $\bm{M}_{T}$ is aligned with the associative matrix ${\bm{M}_{S}}$ with the help of the translator matrix $\bm{M}_F$. Then $\bm{M}_F$ is also used in the alignment between attention matrix $\bm{M}_A$ and ${\bm{M}_{S}}$. After the series alignment and regulation process, we compute the character-level similarity between attention and ${\bm{M}_{S}}$ within the weight regulator in the correction stage (Right). Besides, note that parameters of BERT in the alignment stage and correction are real-time sharing.}
    \label{fig:3}
\end{figure*}

\section{Methodology}
\label{methodology}
\subsection{Construction of Associative Matrix}
\label{akn}

People measure the relations among the characters according to knowledge from experience, which is similar to the statistic logic in AKN. Therefore, based on the associative relations from AKN, we construct the associative matrix $\bm{M}_{S}$ of the characters with a contextification process to represent the interpretable statistic logic of the given sentence (with length $d$), which is defined as:
\begin{equation}
    {\bm{M}_{S}}_{i,j}=\sigma\frac{\bm{A}_{i,j}}{\mathrm{Avg}({\dot{\bm{A}}_{i:}})}-0.5
    \label{eq:2}
\end{equation}
which $\bm{M}_{S} \in {\mathbb{R}^{d\times d}}$, ${\bm{M}_{S}}_{i,j}$ is the associative score of character pair $\langle i,j \rangle$, $\dot{\bm{A}}_{i:}$ is the corresponding the association score among the $i-th$ character and the context sentence, and ${\bm{M}_{S}}_{i:}$ is the $i$-th row of $\bm{M}_{S}$. Besides, $\sigma(\cdot)$ and $\mathrm{Avg}(\cdot)$ is function of sigmoid and average.

\subsection{Quantify BERT Logic as Transforming Matrix}
\label{transforming matrix}
BERT \cite{devlin-etal-2019-bert} consists of embedding layers, attention layers, and linear layers. From the bottom to the top, the hidden layers transform the embedding representation into the semantic representation, where the embedding representation and the semantic representation fit in the same tensor-shape. We define the complex transformation in BERT with the hidden size of $H$ as:
\begin{equation}
    \boldsymbol{F}=\mathrm{Transforming}(\boldsymbol{E})
    \label{eq:3}
\end{equation}
where $\boldsymbol{E} \in{\mathbb{R}^{d\times H}}$ indicates the embedding representation and $\boldsymbol{F} \in{\mathbb{R}^{d\times H}}$ indicates last hidden representation. $\mathrm{Transforming}(\cdot)$ is defined as the transformation process in BERT.

The transformation function $\mathrm{transforming}(\cdot)$, which represents the processing of the BERT layers, is considered as the comprehensive transformation logic of BERT. By use of the least-squares function, we approximately quantify the transformation process as a transforming matrix $\bm{M}_T$:
\begin{equation}
    \bm{M}_{T}=\mathrm{LeastSquares}(\boldsymbol{E},\boldsymbol{F})
    \label{eq:4}
\end{equation}
where $\bm{M}_{T} \in{\mathbb{R}}^{d \times d}$, and $\mathrm{LeastSquares}(\cdot)$ is least-squares function. Transforming matrix $\bm{M}_T$, the solution of least-square equation, serves as approximate quantification of the transforming process from the $\boldsymbol{E}$ to $\boldsymbol{F}$ and represents the uninterpretable BERT logic contrasted to $\bm{M}_S$.

\subsection{Translate BERT Logic to Interpretable Statistic Logic}
\label{translator matrix}
From the perspective of linguistics and experience, the characters that co-occurred in phrases are considered associated. In contrast, the uninterpretable language model conducts the modeling process with irregular influence logic of the representations regardless of the relations among the characters. Even though BERT logic is quantified as a transforming matrix, the uninterpretability still constrains the direct understanding of people. Inspired by the conception of translation, we introduce a translator matrix $\bm{M}_F$, which serves as the translator between the flattened transforming matrix $\bar{\bm{M}}_T$ and associative matrix $\bar{\bm{M}}_S$. We aim to find the appropriate translator matrix to indirectly fit the unattainable alignment between the uninterpretable BERT logic and interpretable statistical logic. Note that the translator matrix $\bm{M}_F$, which is parameter-isolated from the backbone of AxBERT, is trained in parallel with the correction task according to:
\begin{equation}
    L_F=1-S_{T,S}
    \label{eq:7}
\end{equation}
\begin{equation}
    S_{T,S}=\mathrm{CosSim}_{-1}(\bm{M}_F \times \bar{\bm{M}}_T,\bar{\bm{M}}_S)
    \label{eq:6}
\end{equation}
which $\mathrm{CosSim}_{-1}(\cdot)$ is the cosine similarity function at $-1$-th dimension, $\bar{\bm{M}}_T \in {\mathbb{R}}^{d^2 \times 1}$, $\bar{\bm{M}}_S \in {\mathbb{R}}^{d^2 \times 1}$, $\bm{M}_F$ is the translator matrix and $\bm{M}_F \in {\mathbb{R}}^{d^2 \times d^2}$, $L_F$ is the objective for training of the translator matrix. Wherein, the flatten operation for ${\bm{M}}_T$ and ${\bm{M}}_S$ enable the communication among the different semantic representations of the characters.

\subsection{Attention Regulation for Interpretability via Translator Matrix}
\label{attention align}
As we mentioned in section \ref{related_work}, the redundancy distributed in the representations in BERT adversely affects the analysis of BERT. In order to reduce the redundancy in AxBERT, we introduce a regulation on attention to integrating the interpretable statistic logic into BERT. Attention layers, the most significant component of BERT, serve as the key to integrating statistical information. While every component in BERT performs under various logic, even attention is unable to comprehensively represent BERT, but the regulation process enables the progressive unification of the different logic in learning. Besides, because the bottom attention layers are more concerned with character structure \cite{belinkov-etal-2017-neural,jawahar-etal-2019-bert}, a dynamical attention combination method is applied based on the number of attention layers, which realizes a more comprehensive capturing for attention information. The alignment between the flattened attention matrix $\bar{\bm{M}}_A$ and associative matrix $\bar{\bm{M}}_S$ is defined as:
\begin{equation}
    S_{A,S}=\mathrm{CosSim}_{-1}(\bar{\bm{M}}_A,{\bm{M}_F}^{-1} \times \bar{\bm{M}}_S)
    \label{eq:8}
\end{equation}
\begin{equation}
    {\bm{M}}_A=\sum _{i=0} ^{layer} (1-\frac {i}{layer}) \sum_{j=0}^{head}\boldsymbol{AttDis}_{i,j}
    \label{eq:9}
\end{equation}
where $\bar{\bm{M}}_A \in{\mathbb{R}^{d^2 \times 1}}$, ${\bm{M}_F}^{-1}\in{\mathbb{R}^{d^2 \times d^2}}$ and $\boldsymbol{AttDis}_{i,j} \in{\mathbb{R}^{d \times d}}$. $\boldsymbol{AttDis}_{i,j}$ is the attention distribution from the attention head of $i$-th layer and $j$-th head. $layer$ and $head$ are the layer number and attention head number of BERT encoder. 

\subsection{Regulation for Correction Task with Weight Regulator}
\label{weight regulator}

Generally, in BERT-based correction methods, the semantic representations of the characters are modeled based on the context of the given sentence. However, the semantic influence from the irrelative characters or the error characters will decrease the accuracy of the modeling process. In order to maintain an appropriate modeling process in AxBERT, we apply a weight regulator to regulate the attention distributions in the correction stage. Specifically, the errors are distributed with lower scores compared with the other characters in AKN. By aligning the associative matrix with the combined attention matrix ${\bm{M}}_A$ at character level, the sequences of the similarity scores of the given sentence are obtained, where the error positions are computed as lower scores than the other positions. The weight $\boldsymbol{W}$ in the weight regulator is computed as following:
\begin{equation}
    \boldsymbol{W}=\mathrm{CosSim}_{-2}({\bm{M}_A}_{in}+{\bm{M}_A}_{out},\bm{M}_S)
    \label{eq:10}
\end{equation}
\begin{equation}
    {\bm{M}_A}_{in}=\frac {{\bm{M}_A}^T}{\mathrm{Avg_{col}}({\bm{M}_A})^T}
    \label{eq:11}
\end{equation}
\begin{equation}
    {\bm{M}_A}_{out}=\frac {M_A}{\mathrm{Avg_{row}}({\bm{M}_A})}
    \label{eq:12}
\end{equation}
where $\boldsymbol{W} \in \mathbb{R}^{d}$, ${\bm{M}_A}_{in} \in \mathbb{R}^{d}$ and ${\bm{M}_A}_{out} \in \mathbb{R}^{d}$. $\mathrm{CosSim}_{-2}(\cdot)$ is the cosine similarity function at $-2$-th dimension (character-level), $\mathrm{Avg_{row}}(\cdot)$ and $\mathrm{Avg_{col}}(\cdot)$ are row and column average functions in the matrix. Note that different from the undirected associative score, the attention between characters is directed, which contains two kinds of degrees as ${\bm{M}_A}_{in}$ for the in-degree and ${\bm{M}_A}_{out}$ for the out-degree respectively located in columns and rows in extracted attention matrix $\bm{M}_A$. By accumulating ${\bm{M}_A}_{in}$ and ${\bm{M}_A}_{out}$, the attention is transferred into the undirected form, which fits the form of the associative matrix $\bm{M}_S$.


In order to maintain the appropriate semantic influence of characters, based on the obtained character-level weight $\boldsymbol{W}$, we introduce a weight matrix $\bm{M}_W$ to regulate it in the correction stage, which fits the shape of the attention distributions. As a result, the regulated out-degrees of inappropriate characters are decreased, and the out-degrees of appropriate characters are increased. The weight matrix $\bm{M}_W$ in the weight regulator is computed according to:
\begin{equation}
    \bm{M}_W=\boldsymbol{W} \times (\frac{1}{\boldsymbol{W}})^T\cdot \mathrm{diag}(\boldsymbol{W}_1,\dots,\boldsymbol{W}_d)
    \label{eq:13}
\end{equation}
\begin{equation}
    \boldsymbol{AttDis}_{{R_i}:}=(1-\frac{i}{layer})\boldsymbol{AttDis}_i\bm{M}_W
    \label{eq:14}
\end{equation}
where $\bm{M}_W \in \mathbb{R}^{d \times d}$, $\boldsymbol{AttDis}_R$ and $\boldsymbol{AttDis}$ respectively indicates the regulated and original attention distribution, which $\boldsymbol{AttDis}_R \in \mathrm{R}^{d \times d}$ and $\boldsymbol{AttDis} \in \mathrm{R}^{d \times d}$. $\mathrm{diag}(\boldsymbol{W}_1,\dots,\boldsymbol{W}_d)$ is the diagonal matrix value of $\boldsymbol{W}$. Besides, we introduce a weight decay process for the regulation process to decrease the regulation intensity for the attention layers on the top.

\subsection{Training of AxBERT}
The objective $L$ is composed of $L_A$ and $L_C$ corresponding to the alignment and correction stages, which are defined as:
\begin{equation}
    L=\lambda(L_C)+(1-\lambda)(L_A)
    \label{eq:15}
\end{equation}
\begin{equation}
    L_C=-\sum_{t=1}^{T'} \log P(\mathrm{y}_t|\mathrm{X})
    \label{eq:16}
\end{equation}
\begin{equation}
    L_A=1-S_{A,S}
    \label{eq:17}
\end{equation}
where $L_C$ is the objective of the correction task, $L_A$ is the objective of the attention alignment. $S_{A,S}$ defined in equation \ref{eq:8}. $\lambda$ is the combining factor, and we set 0.8 in our training \footnote{The main task in AxBERT is spelling correction, so we preset $\lambda$ from 0.5 to 1.0 and finally set it as 0.8 after experiments.}.

Additionally, we design a pre-train process for AxBERT, where the massive pre-train process is applied to improve the generalization ability of AxBERT. Specifically, We randomly replace 13.5\% of the characters in the correct sentences with the random tokens. For the replaced sentences, we conduct the alignment task and correction task with the same objectives defined in formula \ref{eq:15}-\ref{eq:17}.

\section{Experiments}
\label{experiments}
\subsection{Settings}

The pretrained Chinese BERT is used in AxBERT. We opt streams of 128 tokens, a small batch of size 32, and learning rates of 2e-5 and 4e-5 for the 30-epoch-training of transforming matrix and the correction components. Additionally, the dropout rate of 0.3 is used for the embedding layers, scale-dot product attention, and hidden layers in BERT and ReLU function.

\subsection{Datasets}

\begin{table*}[t]
\centering
\label{table:dataset}
\caption{Statistics information of the used datasets.}
\begin{tabular}{ccccc}
\hline
Dataset  & TrainSet & TestSet & Dataset Level & Usage \\ \hline
SIGHAN15 & 6,526    & 1,100   & Sentence-level & For correction evaluations and interpretable analysis     \\
SIGHAN14 & 2,339    & 1,062   & Sentence-level & For correction evaluations      \\
HybirdSet   & 274,039  & 3,162   & Sentence-level & For training of AxBERT      \\
CLUE     & 2,439    & -       & Article-level & For initialization of AKN      \\ \hline
\end{tabular}
\end{table*}

{\setlength{\parindent}{0cm}
\textbf{SIGHAN} \cite{yu-etal-2014-overview,tseng-etal-2015-introduction}
, a benchmark for Traditional Chinese spelling check evaluation, is used in our training and experiments, which contains SIGHAN14 and SIGHAN15 datasets. We follow the same pre-processing procedure as \cite{wang-etal-2019-confusionset} for SIGHAN to convert the characters to simplified Chinese, which is widely used in the baseline methods. Wherein, SIGHAN14 consists of a 2,339-sentence-TrainSet and a 1,062-sentence-TestSet, and SIGHAN15 consists of a 6,526-sentence-TrainSet and a 1,100-sentence-TestSet. We use the TestSets as the benchmark dataset in our experiment and the TrainSets as part of the training corpus.
}

{\setlength{\parindent}{0cm}
\textbf{HybirdSet} 
\cite{wang-etal-2018-hybrid} is a method for automatic corpus generation for Chinese spelling check, which conducts the OCR- \cite{tong1996statistical} and ASR-based \cite{hartley2005method} methods to generate the visually or phonologically resembled spelling errors. Hybird dataset is composed of a 274,039-sentence-TrainSet and a 3,162-sentence-TestSet. In the previous works, the TrainSet of Hybird datasets is used the training of the correction methods \cite{wang-etal-2019-confusionset,cheng2020spellgcn}. We adopted the same strategy and constructed the training corpus by mixing the TrainSets of SIGHAN and the Hybird.
}

{\setlength{\parindent}{0cm}
\textbf{CLUE} \cite{xu-etal-2020-clue}
CLUE, an open-ended, community-driven project, is the most authoritative natural language understanding benchmark for Chinese including 9 tasks spanning several well-established single-sentence/sentence-pair classification tasks. We use the news dataset in the CLUE to initialize associative knowledge network, which contains 2,439 articles.
}

\subsection{Comparison Approaches}

We evaluate our method of Chinese spelling correction and compare it with several approaches as the baseline. Besides, we also evaluate original BERT masked language model, which is initialized with the same setting as the BERT encoder in AxBERT. The evaluation of original BERT here is introduced to verify the effectiveness of the alignments and regulations in AxBERT, where the strategy is also adopted in previous works \cite{cheng2020spellgcn,liu-etal-2021-plome,nguyen2021domain}. The performance of trained original BERT varies in different works, we include our best effort to obtain the BERT model to complete the experiments.

{\setlength{\parindent}{0cm}
\textbf{HanSpeller++} conducts a multi-stepped reranking strategy by Hidden Markov Language Model for correction task, which is a remarkable rule-based spelling correction method \cite{xiong2015hanspeller}.
}

{\setlength{\parindent}{0cm}
\textbf{Confusionset} introduce the copy strategy into Seq2Seq model for spelling correction task \cite{wang-etal-2019-confusionset}.
}

{\setlength{\parindent}{0cm}
\textbf{SoftMask} is a BERT-based spelling correction method with a soft-mask generator, where the soft-masked strategy is similar to the concept of error detection \cite{zhang2020spelling}.
}

{\setlength{\parindent}{0cm}
\textbf{FASPell} conducted the Seq2Seq prediction by incorporating BERT with additional visual and phonology features \cite{hong2019faspell}.
}

{\setlength{\parindent}{0cm}
\textbf{SpellGCN} incorporated BERT and the graph convolutional network initialized with phonological and visual similarity knowledge for Chinese spelling correction \cite{cheng2020spellgcn}.
}

{\setlength{\parindent}{0cm}
\textbf{PLOME} integrates the phonological and visual similarity knowledge into a pre-trained masked language model with a large pre-train corpus consisted of one million Chinese Wikipedia 
 pages. And it is the SOTA in previous work \cite{liu-etal-2021-plome}.
}

{\setlength{\parindent}{0cm}
\textbf{HeadFilt} is an adaptable filter for Chinese Spell Check, which conducts the domain-shift conditioning problem by introducing a hierarchical embedding according to the pronunciation similarity and morphological similarity \cite{nguyen2021domain}.
}


\subsection{Evaluation Method}

For the evaluation of the spelling correction performance, we use the same evaluation matrix to assess the precision, recall, and F1-score at sentence-level and character-level with the same evaluation matrix of the previous works \cite{wang-etal-2019-confusionset,zhang2020spelling,cheng2020spellgcn}. Note that because of the difference between traditional Chinese and simplified Chinese, the wrong cases in the given converted TestSet is partly incorrect. Therefore, we directly evaluate the result by comparing the sentences in the TestSet with the predicted sentences from AxBERT and the baselines.

In order to exhibit the regulatability and quantitatively analyze of AxBERT. We design the interpretable analysis to evaluate the interpretability of our method on SIGHAN-15 contained two sub-analysis as similarity analysis and controllable analysis. For the similarity analysis, we compute the cosine similarity between the attention and associative distributions. Additionally, the controllable analysis illustrates the using case indicated the situation that the specific character pairs are not expected to be modified in correction. For a given sentence with errors which is able to be successfully corrected, by changing the associative scores among the related character pairs contained the errors, we assess the ratio between the number of retaining errors and the total changed errors.

\subsection{Main Results}

\begin{table*}[t]
\caption{The performance of our method and baseline methods.}
\label{tab:main result}
\begin{center}
\resizebox{\linewidth}{!}{
\renewcommand{\arraystretch}{1}
\begin{tabular}{clcccccclcccccc}
\hline
\multicolumn{2}{c}{\multirow{2}{*}{Method}} & \multicolumn{6}{c}{Sentence Level}                                                            &  & \multicolumn{6}{c}{Character Level}                                                           \\ \cline{3-15} 
\multicolumn{2}{c}{}           & \multicolumn{3}{c}{Detection}                 & \multicolumn{3}{c}{Correction}                &  & \multicolumn{3}{c}{Detection}                 & \multicolumn{3}{c}{Correction}                \\ \hline
\multicolumn{2}{c}{SIGHAN14}                & P             & R             & F1            & P             & R             & F1            &  & P             & R             & F1            & P             & R             & F1            \\ \hline
\multicolumn{2}{c}{ConfusionSet}            & -             & -             & -             & -             & -             & -             &  & 63.2          & 82.5          & 71.6          & 79.3          & 68.9          & 73.7          \\
\multicolumn{2}{c}{FASPell}                 & 61.0          & 53.5          & 57.0          & 59.4          & 52.0          & 55.4          &  & -             & -             & -             & -             & -             & -             \\
\multicolumn{2}{c}{SpellGCN}                & 65.1          & \textbf{69.5} & 67.2          & 63.1          & \textbf{67.2} & 65.3          &  & 83.6          & \textbf{78.6} & 81.0          & 97.2          & \textbf{76.4} & \textbf{85.5} \\
\multicolumn{2}{c}{HeadFilt}                & \textbf{82.5} & 61.6          & 70.5          & 82.1          & 60.2          & 69.4          &  & -             & -             & -             & -             & -             & -             \\
\multicolumn{2}{c}{BERT}                    & 81.6          & 64.1          & 71.8          & 81.0          & 62.6          & 70.6          &  & \textbf{89.4} & 74.1          & 81.0          & 96.9          & 71.8          & 82.5          \\
\multicolumn{2}{c}{AxBERT}                  & 81.9          & 64.4          & \textbf{72.1} & \textbf{81.7} & 63.2          & \textbf{71.2} &  & 87.8          & 76.2          & \textbf{81.6} & \textbf{98.2} & 74.8          & 84.9          \\ \hline
\multicolumn{2}{c}{SIGHAN15}                & P             & R             & F1            & P             & R             & F1            &  & P             & R             & F1            & P             & R             & F1            \\ \hline
\multicolumn{2}{c}{HanSpeller++}            & 80.3          & 53.3          & 64.0          & 79.7          & 51.5          & 62.5          &  & -             & -             & -             & -             & -             & -             \\
\multicolumn{2}{c}{ConfusionSet}            & -             & -             & -             & -             & -             & -             &  & 66.8          & 73.1          & 69.8          & 71.5          & 59.5          & 69.9          \\
\multicolumn{2}{c}{SoftMask}                & 73.7          & 73.2          & 73.5          & 66.7          & 66.2          & 66.4          &  & -             & -             & -             & -             & -             & -             \\
\multicolumn{2}{c}{FASPell}                 & 67.6          & 60.0          & 63.5          & 66.6          & 59.1          & 62.6          &  & -             & -             & -             & -             & -             & -             \\
\multicolumn{2}{c}{SpellGCN}                & 74.8          & 80.7          & 77.7          & 72.1          & 77.7          & 75.9          &  & 88.9          & 87.7          & 88.3          & 95.7          & 83.9          & 89.4          \\
\multicolumn{2}{c}{HeadFilt}                & 84.5          & 71.8          & 77.6          & 84.2          & 70.2          & 76.5          &  & -             & -             & -             & -             & -             & -             \\
\multicolumn{2}{c}{PLOME}                   & 77.4          & \textbf{81.5} & 79.4          & 75.3          & \textbf{79.3} & 77.2          &  & \textbf{94.5} & \textbf{87.4} & \textbf{90.8} & \textbf{97.2} & \textbf{84.3} & \textbf{90.3} \\
\multicolumn{2}{c}{BERT}                    & 84.1          & 75.9          & 79.8          & 83.6          & 72.7          & 77.8          &  & 92.1          & 84.7          & 88.2          & 95.3          & 80.7          & 87.4          \\
\multicolumn{2}{c}{AxBERT}                  & \textbf{88.4} & 78.7          & \textbf{83.3} & \textbf{88.1} & 76.2          & \textbf{81.7} &  & 92.0          & 84.9          & 88.3          & 96.9          & 82.3          & 89.2          \\ \hline
\end{tabular}
}
\end{center}
\end{table*}

The experiment results are illustrated in Table \ref{tab:main result}, while part of the results of the baseline methods is incomparable with other work. P, R, and F1 denote the precision, recall, and F1 score, respectively. 

On the sentence-level evaluation, the F1 score of AxBERT on detection and correction evaluation is advanced compared with the baselines, especially in SIGHAN15, the performance is improved by more than 3 absolute points. And the improvements are also reflected in the precision evaluations. Besides, while AxBERT performs a balance evaluation results in the different indexes, the experimental results of the baseline exhibit the unbalance performance of the baselines in precision and recall, e.g., SpellGCN achieves an advanced performance in recall but is unable to precisely detect and correct the errors. AxBERT is advanced in precision by almost 4 points in SIGHAN15 and first-class in SIGHAN14. And the performance in the recall is also better than most of the methods. From the results, we find that the BERT-based correction methods (including HeadFilt, BERT, AxBERT) are different from the other methods in correction strategy, where AxBERT serves as the SOTA in F1-score. Compared with the other baselines, the higher precision reflects that the AxBERT is "cautious" in correction. We believe that the main reason for the "cautious" tendency is the introduction of the alignment and regulation in AxBERT. The integrated interpretable logic and the additional validation of the character relations from AKN improve the robustness of AxBERT and provide a consistent correction process.

Previous works opted for the sentence-level evaluation as the first choice compared with the character-level evaluation \cite{hong2019faspell,zhang2020spelling,nguyen2021domain}. Even if we think that sentence-level evaluation is more convincing, the result of the character-level evaluation is illustrated. The character-level evaluation results illustrate that AxBERT is not able to achieve the SOTA correction performance, but the performance of AxBERT is the first class compared with the baselines. The SOTA methods, PLOME \cite{liu-etal-2021-plome} on SIGHAN15 and SpellGCN \cite{cheng2020spellgcn} on SIGHAN14, are respectively advance 1.1 and 0.6 points than AxBERT on the two datasets, which both reflects close performance gaps between AxBERT and the SOTA methods. As for the advanced performance of PLOME \cite{liu-etal-2021-plome}, it conducts a massive pretraining on a dataset composed of 162.1 million sentences, which is unachievable for the other methods, including AxBERT and other baselines. Revisiting the methods trained on comparable sizes of dataset, AxBERT obviously achieves attractive performance without any additional features. Definitely, the additional features such as phonic and shape can effectively improve the correction ability of the methods, which is widely used in recent correction methods, including FaSPell \cite{hong2019faspell}, SpellGCN \cite{cheng2020spellgcn} and HeadFlit \cite{nguyen2021domain}. However, the typical additional features, phonic and shape, are commonly useless for the other NLP classification and generation tasks \cite{Zhou2020,Liu2021pre,feng-etal-2021-survey}, which indicates an unsatisfied generalization ability of the methods. Moreover, the single semantics-relied methods, like AxBERT, are designed to enhance the semantic modeling ability of the language models, which are more universal among the NLP tasks compared with the methods integrated with the additional character features. And the different correction strategy also reflects in the character-level evaluation. For the character-level evaluation, AxBERT is better than most of the baselines but still lower than the SOTA, even if there is not a significant gap between them cause that the cautious strategy makes the number of corrected samples of AxBERT less than the other methods.

\subsection{Interpretable Analysis}

\begin{figure*}
    \centering
	\begin{minipage}{0.49\linewidth}
		\centering
        \includegraphics[width=0.95\linewidth]{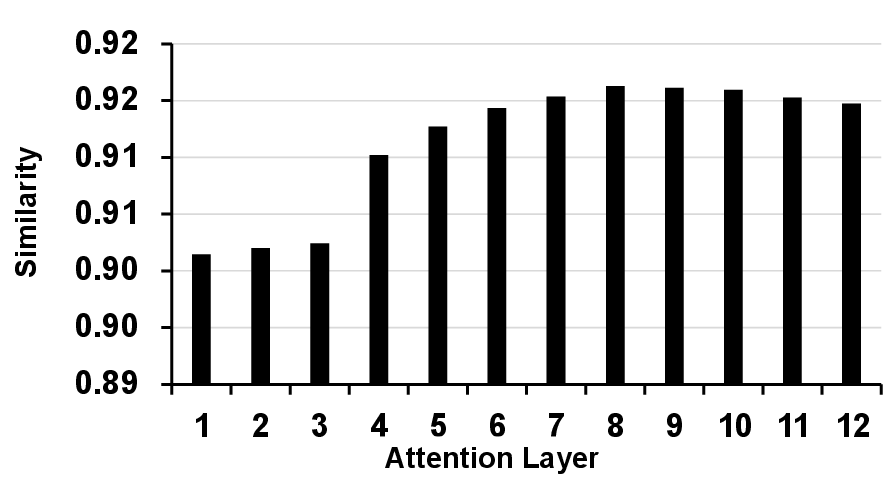}
        \caption{Similarity of association and attention}
        \label{fig:sim}
	\end{minipage}
	\begin{minipage}{0.49\linewidth}
		\centering
        \includegraphics[width=0.95\linewidth]{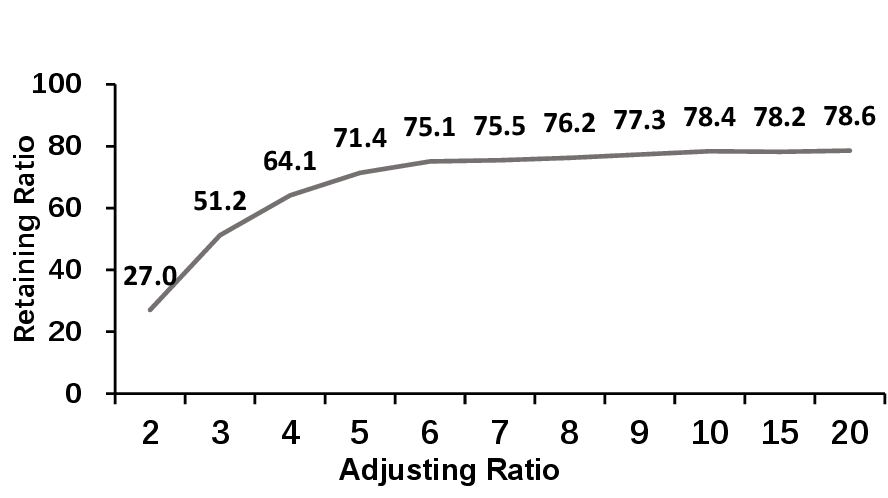}
        \caption{Retaining ratio result}
        \label{fig:control}
	\end{minipage}
\end{figure*}

The data-driven or rule-based machine learning methods, widely considered the root of interpretability, enable a more transparent and consistent decision process \cite{christoph2020Interpretable,verhagen2021two}. The readability of the data-driven methods can directly reflect the transparency in the decision-making process \cite{Zhou2008,jiang-etal-2018-interpretable,Alonso2011}. Moreover, the controllability of the methods is able to significantly demonstrate the consistency of the methods by changing the input data or the presetting rules and evaluating the predictable influence on the output \cite{lee2017making,Tripathy_2020_WACV,TianGGMX19}. Inspired by the characteristics of the interpretable data-driven methods, we analyze the interpretability of AxBERT by (1) comparing the similarities among the statistic data and the hidden distributions and (2) changing the presetting statistic relations to assess the influence on adjusted correction results. The analysis can evaluate the interpretability with the readability of the hidden distributions and the consistency of the model structure from the bottom (input) to the top (output).

The detailed similarities between the associative matrix and the attention distributions are presented in figure \ref{fig:sim}. Figure \ref{fig:sim} depict an extremely high situation in similarity between the hidden distributions and statistic data, which demonstrates the readability of AxBERT. Besides, as the results in the previous works, the low layers prefer to learn the word structure while the top layers prefer to learn the word meanings \cite{belinkov-etal-2017-neural,jawahar-etal-2019-bert}. The similarity results supported the above conclusion. While associative relations are regarded as character relations based on semantics (meaning), the higher layer in AxBERT is assessed as more similar to the associative matrix.

In the controllable analysis, for the successfully corrected errors, we multiple different adjusting ratios to the corresponding associative scores to implicitly influence the weight regulator according to formula \ref{eq:10}, which is designed to simulate the actual application scenarios when people need to modify the correction rules in some specific domains. The errors are expected to keep retained, and we calculate the retaining ratio of adjusted errors to the total. The retaining ratios with the different adjusting ratios are shown in figure \ref{fig:control}, with the increase of the adjusting ratio, the number of retaining errors is increasing. The most rapid increase occurs when the adjusting ratio is set from 2 to 6; after that, the growth rate slowed down and stabilized at around 78\%. We think that the errors that were not successfully retained at last are so fatal in semantics so that the correction method has to handle them to keep the semantic fluency.


\subsection{Case Study}

\begin{table*}
\caption{Correction result of the case study}
\label{tab:case}
\begin{center}
\renewcommand{\arraystretch}{1.2}
\resizebox{0.8\linewidth}{!}{
\begin{tabular}{cc}
\hline
Wrong Sentence     & \begin{CJK}{UTF8}{gbsn}我跟我朋\end{CJK}{\color[HTML]{FE0000} \begin{CJK}{UTF8}{gbsn}唷\end{CJK}}\begin{CJK}{UTF8}{gbsn}打算去法国玩儿\end{CJK}/I plan to travel to France with my fri{\color[HTML]{FE0000}\begin{CJK}{UTF8}{gbsn}***\end{CJK}} \\\hline
Predicted Sentence & \begin{CJK}{UTF8}{gbsn}我跟我朋\end{CJK}{\color[HTML]{009901} \begin{CJK}{UTF8}{gbsn}友\end{CJK}}\begin{CJK}{UTF8}{gbsn}打算去法国玩儿\end{CJK}/I plan to travel to France with my fri{\color[HTML]{009901}end}                        \\\hline
Correct Sentence   & \begin{CJK}{UTF8}{gbsn}我跟我朋\end{CJK}{\color[HTML]{009901} \begin{CJK}{UTF8}{gbsn}友\end{CJK}}\begin{CJK}{UTF8}{gbsn}打算去法国玩儿\end{CJK}/I plan to travel to France with my fri{\color[HTML]{009901}end} \\\hline
Regulated Sentence  & \begin{CJK}{UTF8}{gbsn}我跟我朋\end{CJK}{\color[HTML]{3166FF} \begin{CJK}{UTF8}{gbsn}唷\end{CJK}}\begin{CJK}{UTF8}{gbsn}打算去法国玩儿\end{CJK}/I plan to travel to France with my fri{\color[HTML]{3166FF}***} \\\hline

\end{tabular}
}
\end{center}
\end{table*}

\begin{figure*}[t]
    \centering
    \subfloat[]{\label{fig:case_oa}
    \includegraphics[width=0.33\linewidth, height=3cm]{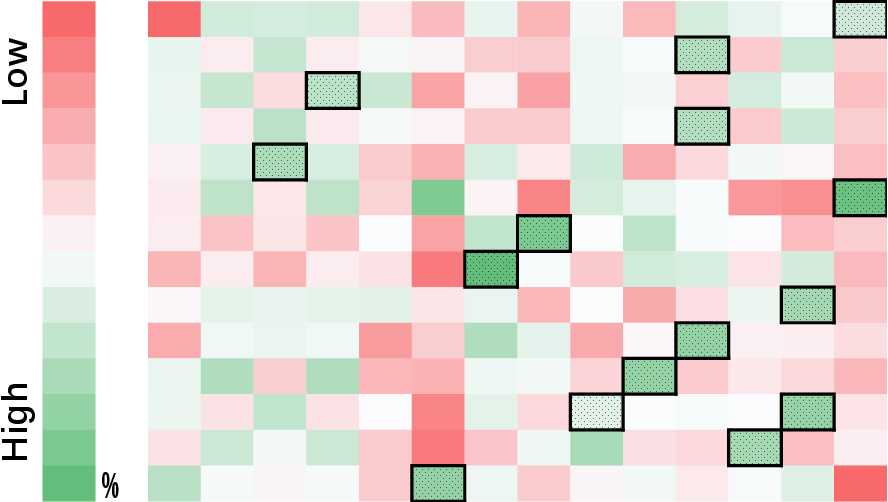}}
    \hspace{1mm}
    \subfloat[]{\label{fig:case_ea}
    \includegraphics[width=0.32\linewidth, height=3cm]{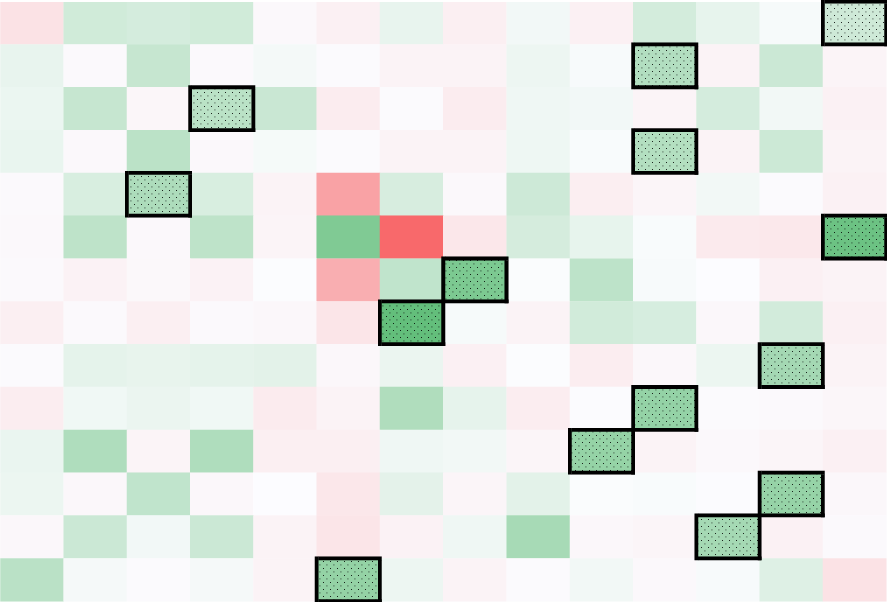}}
    \hspace{1mm}
    \subfloat[]{\label{fig:case_att}
    \includegraphics[width=0.32\linewidth, height=3cm]{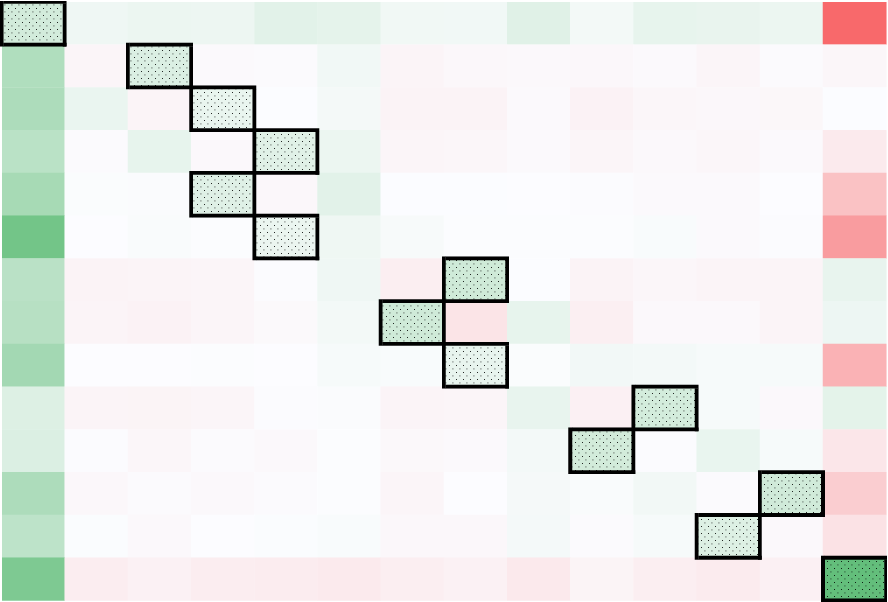}}\\
    \subfloat[]{\label{fig:case_ow}
    \includegraphics[width=0.32\linewidth, height=3cm]{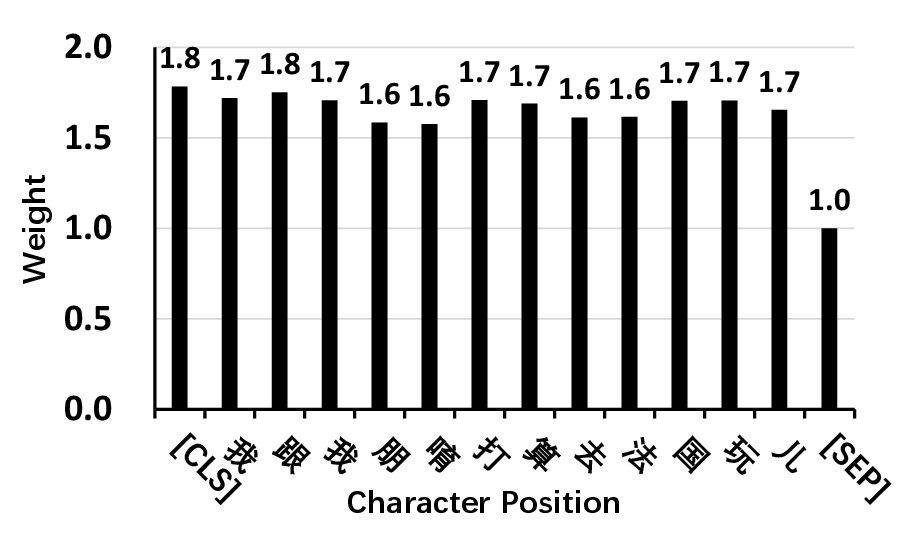}}
    \hspace{1mm}
    \subfloat[]{\label{fig:case_ew}
    \includegraphics[width=0.32\linewidth, height=3cm]{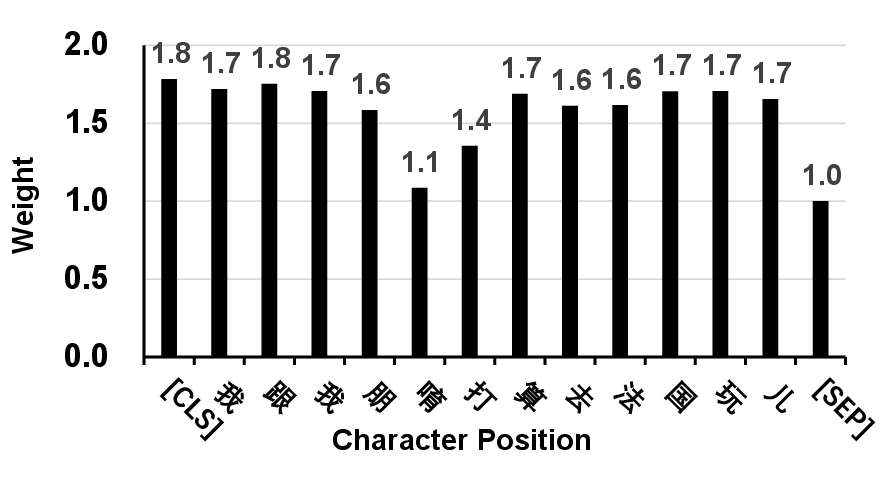}}
    \hspace{1mm}
    \subfloat[]{\label{fig:case_sim}
    \includegraphics[width=0.32\linewidth, height=3cm]{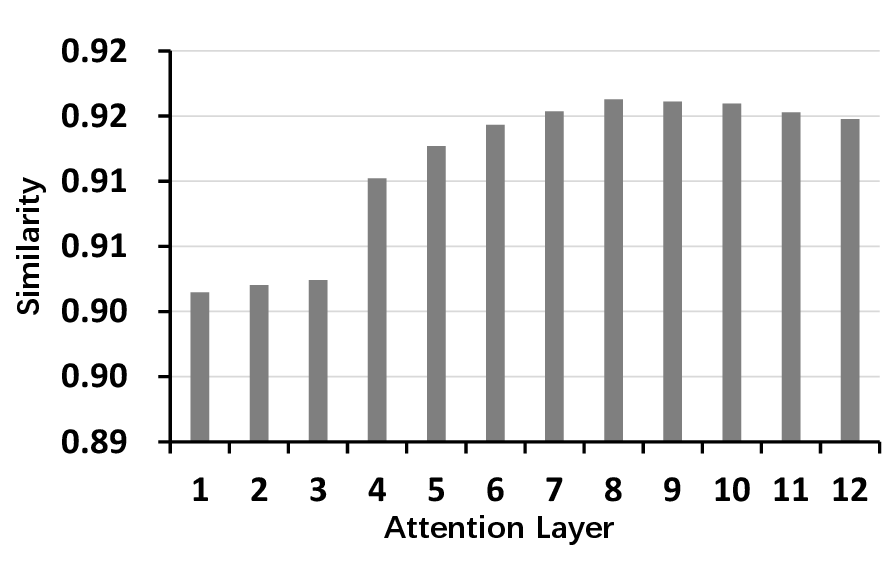}}
    \caption{Result of the case study. (a) Original association. (b) Enhancing association. (c) Attention matrix. (d) Original weight. (e) Enhancing weight. (f) Attention similarities.}
    \label{fig:case_study}
\end{figure*}

As shown in table \ref{tab:case}, the errors \begin{CJK}{UTF8}{gbsn}朋唷/fri***\end{CJK} is successfully corrected to \begin{CJK}{UTF8}{gbsn}朋友/friend\end{CJK}. Besides, we also present the regulated sentence after we adjust the associative relation among \begin{CJK}{UTF8}{gbsn}朋 and 唷\end{CJK}. The associative and attention distributions of the sentence are respectively presented in figure \ref{fig:case_study}\subref{fig:case_oa}, \ref{fig:case_study}\subref{fig:case_ea} and \ref{fig:case_study}\subref{fig:case_att}, where the black-bordered units are the highest score in row. The distributions reflect the relations among the characters. By comparing figure \ref{fig:case_study}\subref{fig:case_oa}, \ref{fig:case_study}\subref{fig:case_ea} and \ref{fig:case_study}\subref{fig:case_att}, green we can discover that the positions of the black-bordered units are distributed similarly, which demonstrate the alignment between attention and AKN. Besides, we also show the weight in weight regulator of the given sentence in figure \ref{fig:case_study}\subref{fig:case_ow}, \ref{fig:case_study}\subref{fig:case_ew}. The corresponding weight of the adjusted characters \begin{CJK}{UTF8}{gbsn}朋 and 唷\end{CJK} are decreased to make them less influence on other characters to keep retained.

\section{Discussion and Conclusion}
This paper reports an interpretable Chinese spelling correction method named AxBERT, which is driven by semantic alignment and regulation. While the alignment and regulation are applied to hidden layers of BERT, the various logic from the components in BERT are unified with clearer semantic relations. Besides, the weight regulator is introduced to regulate the attention distribution to model the sentence in a more appropriate way, which effectively improves the correction performance. In the evaluation of SIGHAN dataset, the effectiveness and interpretability of AxBERT are demonstrated.

With respect to the better generalization ability of our method, AxBERT is also able to be employed in other languages by initializing AKN in different languages for spelling correction. Furthermore, the other spelling correction methods are primarily limited in correction tasks, which is caused by the wide application of the additional character features, and these features are not commonly used in other NLP tasks. Contrarily, the process of semantic regulation and alignment in AxBERT is able to be applied in other NLP tasks to enhance interpretability and semantic modeling ability.

Benefits from interpretability and high performance, AxBERT is able to be widely employed in various usage scenarios. Specifically, after the training with the general correction corpus, simply adjusting for the associative relations enable AxBERT to fit the unusual character correlations in specific domains, such as the medical, biology, and legal domains. In the future, we plan to extend the interpretable information to the decoder structure as a non-autoregression prediction, which can realize a variable-length correction framework in order to tackle wider correction situations.


\bibliography{custom,anthology}
\bibliographystyle{IEEEtran}



\vspace{11pt}

\vfill

\end{document}